\documentclass[lettersize,journal]{IEEEtran}
\usepackage{amsmath,amsfonts}
\usepackage{algorithmic}
\usepackage{algorithm}
\usepackage{array}
\usepackage[caption=false,font=normalsize,labelfont=sf,textfont=sf]{subfig}
\usepackage{textcomp}
\usepackage{stfloats}
\usepackage{hyperref}
\usepackage{url}
\usepackage{verbatim}
\usepackage{graphicx}
\usepackage{cite}
\usepackage{tabularray}
\usepackage{pifont}
\usepackage{multirow}
\usepackage{float}
\usepackage[table,xcdraw]{xcolor}
\usepackage{colortbl}
\definecolor{mycolor1}{RGB}{255,245,157} 
\definecolor{mycolor2}{RGB}{200,230,201} 
\arrayrulecolor[HTML]{000000}  
\usepackage{makecell}
\usepackage{color}
\usepackage{amssymb}
\usepackage{diagbox}
\usepackage{listings}
\begin{document}

\title{\textit{TaleFrame}: An Interactive Story Generation System with Fine-Grained Control and Large Language Models}

\author{
Yunchao Wang$^1$, 
Guodao Sun$^{1,2}$,
Zihang Fu$^1$,
Zhehao Liu$^1$,
Kaixing Du$^1$,
\\ Haidong Gao$^1$,
and Ronghua Liang$^1$,~\IEEEmembership{Senior Member,~IEEE}
\thanks{The authors are with (1) the College of Computer Science and Technology, Zhejiang University of Technology, Hangzhou 310023, China; (2) Zhejiang Key Laboratory of Visual Information Intelligent Processing, Hangzhou 310023, China. (email: wyctears@gmail.com; guodao@zjut.edu.cn; fuzihang@zjut.edu.cn; zhehaoliu.leo@gmail.com; kaixingdu@qq.com; gaohaidong@zjut.edu.cn; rhliang@zjut.edu.cn)}
}

\markboth{Journal of \LaTeX\ Class Files,~Vol.~14, No.~8, August~2021}%
{Shell \MakeLowercase{\textit{et al.}}: A Sample Article Using IEEEtran.cls for IEEE Journals}


\maketitle

\begin{abstract}
With the advancement of natural language generation (NLG) technologies, creative story generation systems have gained increasing attention. However, current systems often fail to accurately translate user intent into satisfactory story outputs due to a lack of fine-grained control and unclear input specifications, limiting their applicability. To address this, we propose TaleFrame, a system that combines large language models (LLMs) with human-computer interaction (HCI) to generate stories through structured information, enabling precise control over the generation process. The innovation of TaleFrame lies in decomposing the story structure into four basic units: entities, events, relationships, and story outline. We leverage the Tinystories dataset, parsing and constructing a preference dataset consisting of 9,851 JSON-formatted entries, which is then used to fine-tune a local Llama model. By employing this JSON2Story approach, structured data is transformed into coherent stories. TaleFrame also offers an intuitive interface that supports users in creating and editing entities and events and generates stories through the structured framework. Users can control these units through simple interactions (e.g., drag-and-drop, attach, and connect), thus influencing the details and progression of the story. The generated stories can be evaluated across seven dimensions (e.g., creativity, structural integrity), with the system providing suggestions for refinement based on these evaluations. Users can iteratively adjust the story until a satisfactory result is achieved. Finally, we conduct quantitative evaluation and user studies that demonstrate the usefulness of TaleFrame. Dataset available at \href{https://huggingface.co/datasets/guodaosun/tale-frame}{https://huggingface.co/datasets/guodaosun/tale-frame}.
\end{abstract}


\section{Introduction}
\IEEEPARstart{T}{he} rapid development of artificial intelligence (AI) has driven significant advancements in natural language processing (NLP), with story generation emerging as a prominent research area. Story generation seeks to create coherent, engaging, and contextually appropriate stories based on input \cite{alhussain2021automatic}. However, existing systems still face several challenges, particularly in consistency, entity development, and control over story progression \cite{wang2023open}. Specifically, translating user intent into the desired story content while maintaining control over plot progression, entity actions, and story structure remains a critical issue \cite{spangher2022sequentially}. Thus, fine-grained control is a key challenge—avoiding logical inconsistencies in generated content and ensuring alignment with user expectations to preserve both plausibility and creativity~\cite{yilmaz2024generative}.

The application of LLMs in recent years has opened up new possibilities for automated story generation. Models such as OpenAI’s GPT-4 and Meta’s Llama-3~\cite{Metallama32024} have demonstrated powerful capabilities in text generation tasks, producing high-quality textual content. However, since LLMs are primarily pre-trained on large-scale, general-purpose data, which limits their ability to generate sufficiently controlled and structured story content. This is particularly evident in tasks that require the generation of specific narrative structures, where LLMs struggle to meet users’ dual requirements for plot details and overall story framework \cite{prabhumoye2020exploring}. As such, improving the control capabilities of LLMs to ensure that generated stories align with user expectations while remaining creative has become an urgent issue to address.

The combination of fine-tuning techniques and structured input has emerged as a promising solution. By fine-tuning LLMs and incorporating structured input prompts, models can be finely adjusted in the story generation domain to produce more structured and user-expected content \cite{wang2024talk,white2023prompt,parthasarathy2024ultimate}. This approach not only improves the quality of generated content but also enhances the stability and reliability of the model in complex story generation tasks. The integration of fine-tuning and structured input clearly provides a new pathway to improving the predictability and consistency of the generated content, addressing the control issues inherent in the generation process.

\begin{figure*}[htbp]
  \centering
  \includegraphics[width=\linewidth]{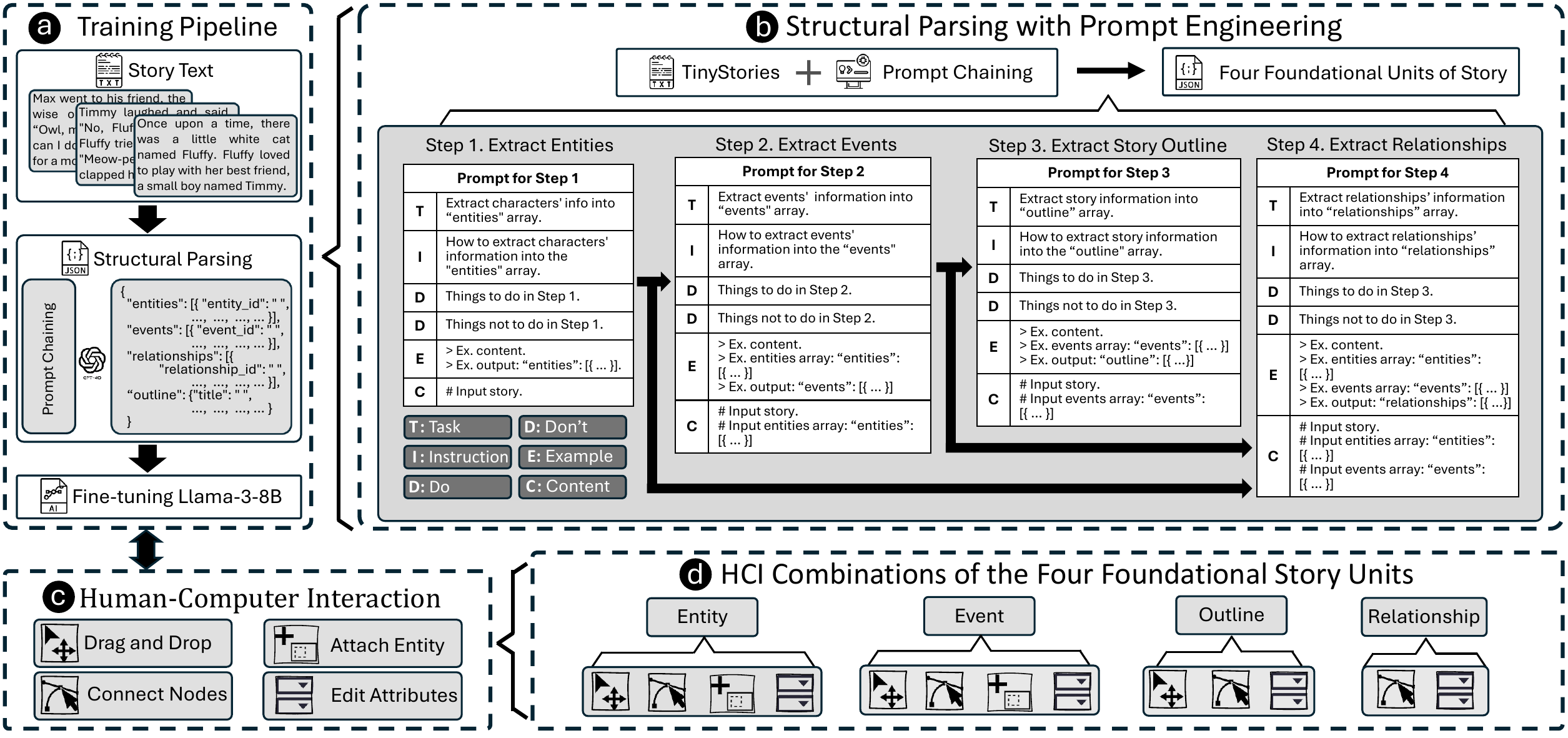}
  \caption{The overview of TaleFrame. (a) The Training pipeline comprises three stages: input story text, story structural parsing, and fine-tuning Llama-3-8B. (b) We decompose the story structure parsing with prompt chaining into four steps: extract entities, extract events, extract story outline, and extract relationships. We summarize the four types of human-computer interaction in TaleFrame (c) and map them to the four foundational units (d).}
  \label{fig:pipeline}
\end{figure*}

In this context, we propose \textit{TaleFrame}, an innovative automated story generation system. TaleFrame decomposes stories into four foundational units: entities, events, relationships, and story outlines, with an intuitive interface that allows users to freely customize the content. This effectively addresses the fine-grained control challenges encountered by existing LLMs in story generation. To improve fine-tuning, we constructed a preference dataset with 9,851 JSON entries to fine-tune the Llama-3-8B model, ensuring the generated stories meet user-specific needs~\cite{hoque2024natural}. Additionally, TaleFrame integrates fine-tuned LLMs with human-computer interaction techniques, ensuring consistency and creativity while optimizing privacy and personalization. This guarantees the diversity and richness of the generated content, offering users greater control over the story creation process.
Figure~\ref{fig:pipeline} is the overview of TaleFrame.
Specifically, the main contributions are as follows:
\begin{itemize}
    \item We propose TaleFrame, an interactive system that generates stories from structured information using fine-tuned LLMs, offering fine-grained control over story generation by translating user intentions into structured story units.
    \item We construct a preference dataset in JSON format with 9,851 entries for fine-tuning LLMs, and provide valuable resources for the visualization and HCI communities.
    \item We conducted an ablation study to validate the four foundational units. We also conducted a user study to validate the usability and effectiveness of TaleFrame for controllable story generation.
\end{itemize}

\section{Related Work}

\subsection{Story writing assistants}
Story writing assistants~\cite{shaikh2023artificially} are software tools that help authors with ideation, plot construction, and narrative development~\cite{storyprism2024, storyassist2024, squibler2024}. 
These tools offer various functions, such as text generation, plot planning, and entity development, aimed at enhancing efficiency and sparking creativity. Some assist in organizing and refining narrative structure~\cite{dramatica2024}, while others generate content autonomously or through crowdsourcing~\cite{huang2020heteroglossia, feldman2021we, beguvs2024experimental}. 
In autonomous generation, computers learn user-defined rules and objectives to make decisions~\cite{lebowitz1984creating, riedl2010narrative, mirowski2023co}, and can simulate reasoning based on roles and case studies~\cite{li2024naruto, chandra2024storytelling}. 
By inputting parameters like theme or outline, users can receive fully AI-generated stories or even complete books~\cite{sudowrite2024, magickpen2024}. 
To give users more control, language models are often guided using sentences or keywords~\cite{clark2018creative, goldfarb2019plan, sun2021iga}. 
With the rise of LLMs, conversational agents are making story refinement accessible to a wider audience~\cite{singh2023hide, rezwana2023designing}. 
Users can also manipulate visual elements to influence story generation~\cite{chung2022talebrush, kim2023cells}. 
However, professional authors have expressed concerns about these tools~\cite{biermann2022tool}. While they can expand on an author’s themes and outlines, they often limit control over specific events or detailed entity interactions~\cite{ozmen2023six, zhang2023visar, draxler2024ai}.

In TaleFrame, we divide the story into four foundational units: \textit{Entity}, \textit{Event}, \textit{Relationship}, and \textit{Outline}. We design a JSON format for each of units to make them editable blocks~\cite{gathani2022grammar}. 
Users can give these blocks properties by \textit{Drag and drop}, \textit{Connect}, \textit{Attach}, and \textit{Edit}.
The strict JSON format allows for better control over the generated stories, making the co-creation process human-driven rather than AI.

\subsection{Prompt engineering}
Prompt engineering is crucial in fine-tuning locally deployed task-specific LLMs~\cite{liu2023pre}. This not only improves data generation efficiency but also reduces the need for manual intervention, saving time and costs. This efficiency is vital for quickly adapting models to market and technological changes.Additionally, diverse techniques and strategies in prompt engineering allow researchers to optimize the performance of LLMs in generating training data, thereby improving fine-tuning efficiency and data quality. 
In story generation, prompt engineering helps create coherent narratives by guiding the model through plot development, entity motivations, and contextual consistency. Techniques like Chain of Thought (CoT) prompting~\cite{shao2023synthetic} allow the model to reason through the narrative, ensuring logical progression. Prompt chaining~\cite{sun2024prompt} breaks complex tasks into smaller steps, supporting the generation of connected events in a story. These methods enhance both data quality and model utility, enabling LLMs to generate more creative and contextually rich content.

We experiment with various prompt strategies during the data generation process, ultimately selecting a prompt-chaining approach integrated with the TIDD-EC framework. In Section 4.2, we detail the parsing of short stories into JSON format and evaluate the performance of four prompt strategies.

\subsection{Story generation with LLMs}
Generating coherent and engaging stories is an important task in the field of AI.
Early research focused on using LLMs, like GPT~\cite{openaio12024}, to generate creative textual content. 
These models typically gain language comprehension and generation capabilities through pre-training on large-scale textual data, followed by fine-tuning for specific story generation tasks~\cite{dong2022survey}. 
For example, models can be trained to learn specific literary styles, narrative structures, or generate stories around a particular theme.
One of the key challenges in story generation is maintaining coherence and consistency in the narrative, including the logical flow of events, entity development, and plot progression~\cite{sevastjanova2022lmfingerprints}. 
Researchers have addressed these issues through techniques such as attention mechanisms~\cite{hu2020introductory} to pre-define outlines, and external knowledge bases to enrich story content and background. 
Fine-tuning further enhances story generation by tailoring models to specific genres or application scenarios, such as education and entertainment~\cite{lin2024data, qin2024empirical}. 
Locally deployed LLMs have also gained traction, especially in resource-limited environments, due to their ability to operate on local devices, reducing latency and enhancing data privacy~\cite{fu2024serverlessllm}. 
By integrating structured datasets and fine-tuning, LLMs effectively address diverse requirements while maintaining performance even under low-resource conditions.

In TaleFrame, we deploy the Llama-3-8B model on an RTX A6000 GPU. To improve the quality of generated stories, we fine-tuned this model using a preference dataset~\cite{rafailov2024direct}. Section 4.2 provides detailed technical descriptions of the fine-tuning process. We also perform an ablation study to compare the performance with different fine-tuning strategies.

\section{Story Unit Identification and Practice Exploration}
In this section, we introduce the preliminary work required to implement TaleFrame. We invited 2 domain experts (E1-E2) and 17 volunteers (P1-P17) to participate. Details of the experts and volunteers are shown in Table~\ref{tab:t1}.

\renewcommand{\arraystretch}{1.2}
\begin{table}[htbp]
\caption{Volunteer Details for TaleFrame}
\label{tab:t1}
\centering
\begin{tabular}{ccccc}
\hline
\multicolumn{1}{c}{Volunteer} & Identity    & Gender     & Age     \\ 
\hline
E1                               & Professional writer             & Male       & 54   \\
E2                               & Doctoral student in Literature  & Female     & 29  \\
P1                               & Primary school student        & Male       & 12   \\
P2                               & College student               & Male       & 20  \\
P3                               & High school student           & Female     & 16  \\
P4                               & College student               & Male       & 19  \\
P5                               & Primary school student        & Male       & 12   \\
P6                               & Primary school student        & Female     & 11  \\
P7                               & College student               & Female     & 21   \\
P8                               & College student               & Male       & 20  \\
P9                               & High school student           & Female     & 17  \\
P10                              & College student               & Male       & 20  \\
P11                              & College student               & Male       & 21  \\
P12                              & High school student           & Male       & 17   \\
P13                              & High school student           & Female     & 17  \\
P14                              & Primary school student        & Male       & 12   \\
P15                              & College student               & Male       & 20  \\
P16                              & College student               & Female     & 21  \\
P17                              & College student               & Male       & 23  \\
\hline
\end{tabular}
\end{table}

\subsection{Interview-Driven Units Identification}
We conducted separate face-to-face semi-structured interviews with E1 and E2 to gain an in-depth understanding of the foundational units that constitute a story. 
The scope of the interviews encompassed the following areas of discussion:
\begin{itemize}
    \item What is your perspective on the structure of stories?
    \item Which foundational units within a story structure are both crucial and relatively independent?
    \item How do you conceptualize and arrange these foundational units when developing new stories?
\end{itemize}

\renewcommand{\arraystretch}{1.2}
\begin{table*}[htbp]
\caption{Definitions and attributes of four foundational units of story}
\label{tab:t2}
\centering
\begin{tabular}{cp{5.7cm}p{9cm}}
\hline
Units        & \multicolumn{1}{c}{Definition}                                                        
& \multicolumn{1}{c}{Attributes}                                                                                         \\ \hline
Entity       & The individuals or entities who play roles within the story.                          
& entity\_id; entity\_name; entity\_identity; entity\_motivation; personality\_traits                  \\
Event        & An occurrence or action that takes place within the story.                            
& event\_id; event\_time; event\_location; event\_details; event\_importance; earlier\_event; later\_event\\
Relationship & The connections or modes of interaction between the entities.                         
& relationship\_id; included\_entities; emotional\_type; action\_type; action\_direction; relationship\_strength; relationship\_evolution \\
Outline        & A high-level framework that organizes the main events and entities, providing a structured overview of the story’s flow.
& title; story\_description; story\_structure (\textit{beginning}, \textit{middle}, \textit{climax}, \textit{ending})\\ \hline
\end{tabular}
\end{table*}

E1 is an experienced professional writer. In his view, events and entities are the two core units, which contribute individually and in tandem to the development of the story. 
\textit{``In `An Introduction to the Structural Analysis of Narrative'~\cite{barthes1975introduction}, Barthes delineates three levels of story structure: functions, actions, and narration. Function and action are the most intuitive when initially constructing a story. Function relates to the events in the story, while action relates to the behaviors and settings of the entities in the story.''}
When conceptualizing a story, users can start with an event as the starting point, and then shape the traits and development of the entities associated with the event; or they can take an entity as the centerpiece and pull out a series of events surrounding that entity~\cite{trabasso1985causal}.

E2 is a PhD student in literature. She believes that stories can be broken down into the following units: story outline, sequences, propositions, and parts of speech. By progressively combining these structural units in reverse order, people can ultimately make a coherent story~\cite{bal2009narratology}.
\textit{``The parts of speech (such as a person, an action, or an attribute) can form a proposition. Several propositions can be organized into a sequence, and two or more sequences can ultimately constitute a complete narrative.''}
However, mere concatenation of these units does not yield coherent stories. Accordingly, the elements that bind these units together function as distinct structural components, defined as relationships~\cite{stein2012building}. 

Analysis of semi-structured interviews with two domain experts revealed four foundational constructs: entity, event, relationship, and outline. These four units and their definitions align with ones proposed in previous work~\cite{seufert2016espresso,maggini2019learning,oza2021entities,wang2024e2}. 
The specific definitions of these units are detailed in Table~\ref{tab:t2}. In discussions with the experts, we examined the attributes of each foundational unit, noting that each encompasses multiple interpretations. The foundational units of story structure are events; however, the constituent attributes of these events can be implicitly present in the narrative representation. Events possess a degree of relative autonomy within the narrative, and their interactional patterns construct the overall narrative tension and meaning. 
Similarly, the identity of an entity and the content of a relationship can influence the events within a story. However, the specific attributes of an entity’s identity and the content of a relationship can also be independently categorized within their respective units. 

\subsection{Workshop-Driven Practices Exploration}
We recruited 17 volunteers (P1–P17) through school posters and random invitations on the street. The study was approved by the Institutional Review Board (IRB), and all underage volunteers provided guardian consent. All interview materials were securely stored, and no personally identifiable information was involved.
Most volunteers were familiar with creating short stories, with an average self-reported score of 3.8 on a 5-point Likert scale. All volunteers had prior experience in writing stories, 
both by hand or with office software.
Volunteers were divided into groups of approximately four, each accompanied by a staff member. The workshop lasted around 2.5 hours. During the session, volunteers were encouraged to verbalize their thought processes using a ``think-aloud'' approach, which enabled us to capture and analyze both their initial and final outputs, as well as the steps of their creative process.
Additionally, we conducted follow-up interviews with four volunteers. 
By analyzing the workshop recordings and interview data, we identified and categorized four distinct modes of interaction observed during the story creation process. 
The workshop flow is outlined as follows:
\begin{enumerate}
    \item \textbf{Introduction and Overview (15 minutes).} 
    The staff will provide a succinct overview of the workshop's objectives and structure, highlighting the application of the four foundational units through illustrative examples. Volunteers will be reminded to adopt a ``think-aloud'' approach throughout the creative process, ensuring transparency and reflection on their decision-making.
    \item \textbf{Group Discussion and Story Framework Creation (60 minutes).} 
    The staff will distribute cards to volunteers. Each group will engage in a collaborative discussion to develop their own story framework, utilizing the four foundational units. Throughout the process, volunteers will be encouraged to articulate their thoughts and decision-making processes in time. The staff will circulate among the groups, offering timely guidance and support as necessary.
    \item \textbf{Story Sharing and Feedback (45 minutes).} Volunteers will detail their utilization of the four foundational units in story construction. Volunteers will receive formative feedback from both peer and staff, emphasizing the creativity and structural coherence of their stories.
    \item \textbf{Informal Interviews with Volunteers (30 minutes).} The staff will select four volunteers (P1-P4) for informal interviews to explore their creative experiences. The interviews will focus on the following questions: \textcircled{1} Which aspects of your thinking are the most challenging to articulate during the creative process? \textcircled{2} Which of the four foundational units is the most helpful in facilitating your creative process? \textcircled{3} Do you encounter any challenges, and how to address them? \textcircled{4} Do you make any new insights during the creative process?
\end{enumerate}

\begin{figure}[htbp]
  \centering
  \includegraphics[width=\linewidth]{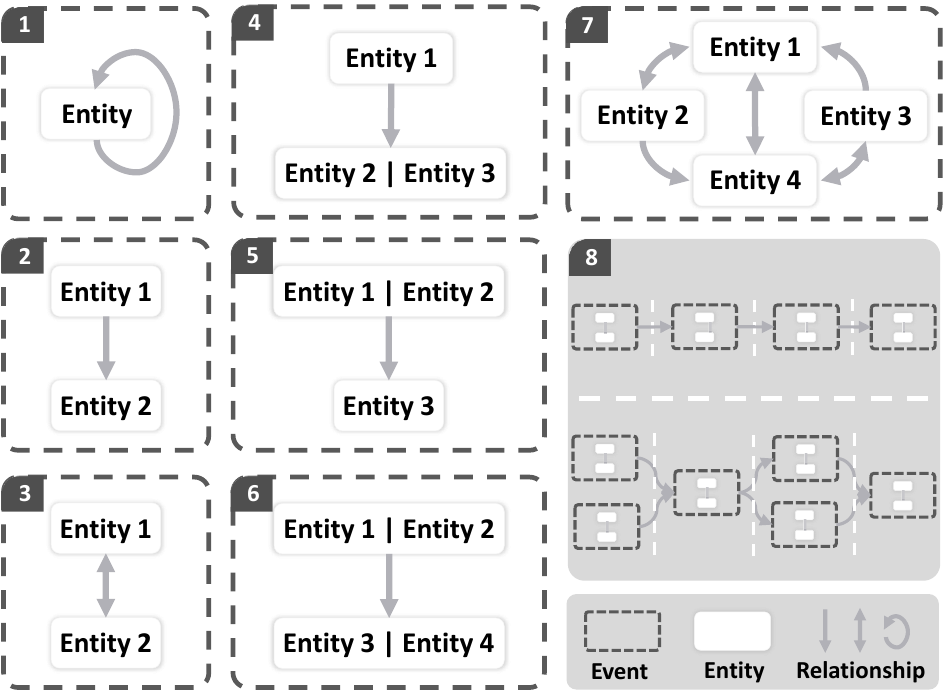}
  \caption{Interactive patterns. This diagram includes a legend at the bottom right, which explains the symbols used throughout the diagram:
        \raisebox{0mm}{\includegraphics[scale=0.01]{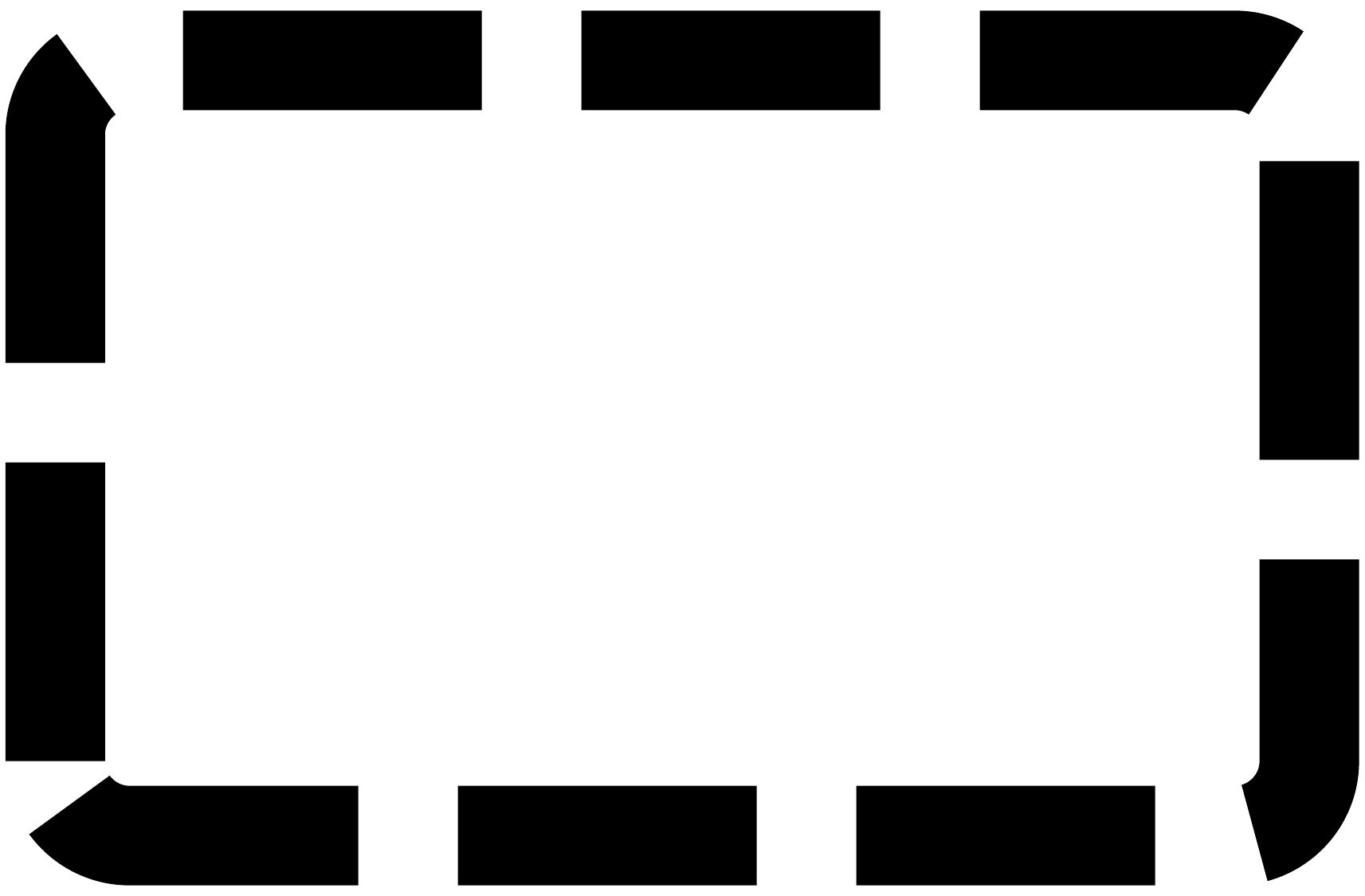}} Event, 
        \raisebox{0mm}{\includegraphics[scale=0.01]{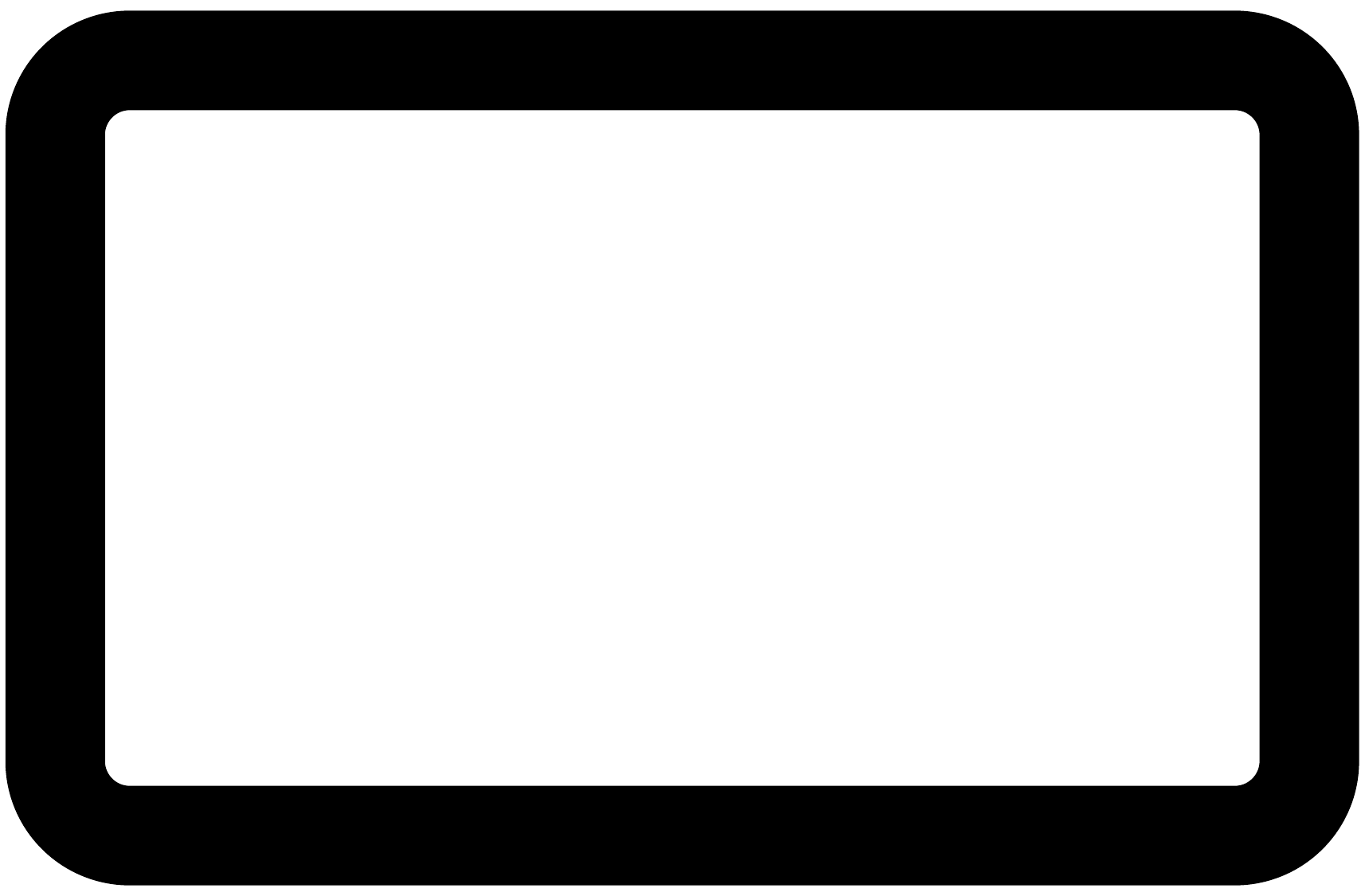}} Entity,
        \raisebox{0mm}{\includegraphics[scale=0.01]{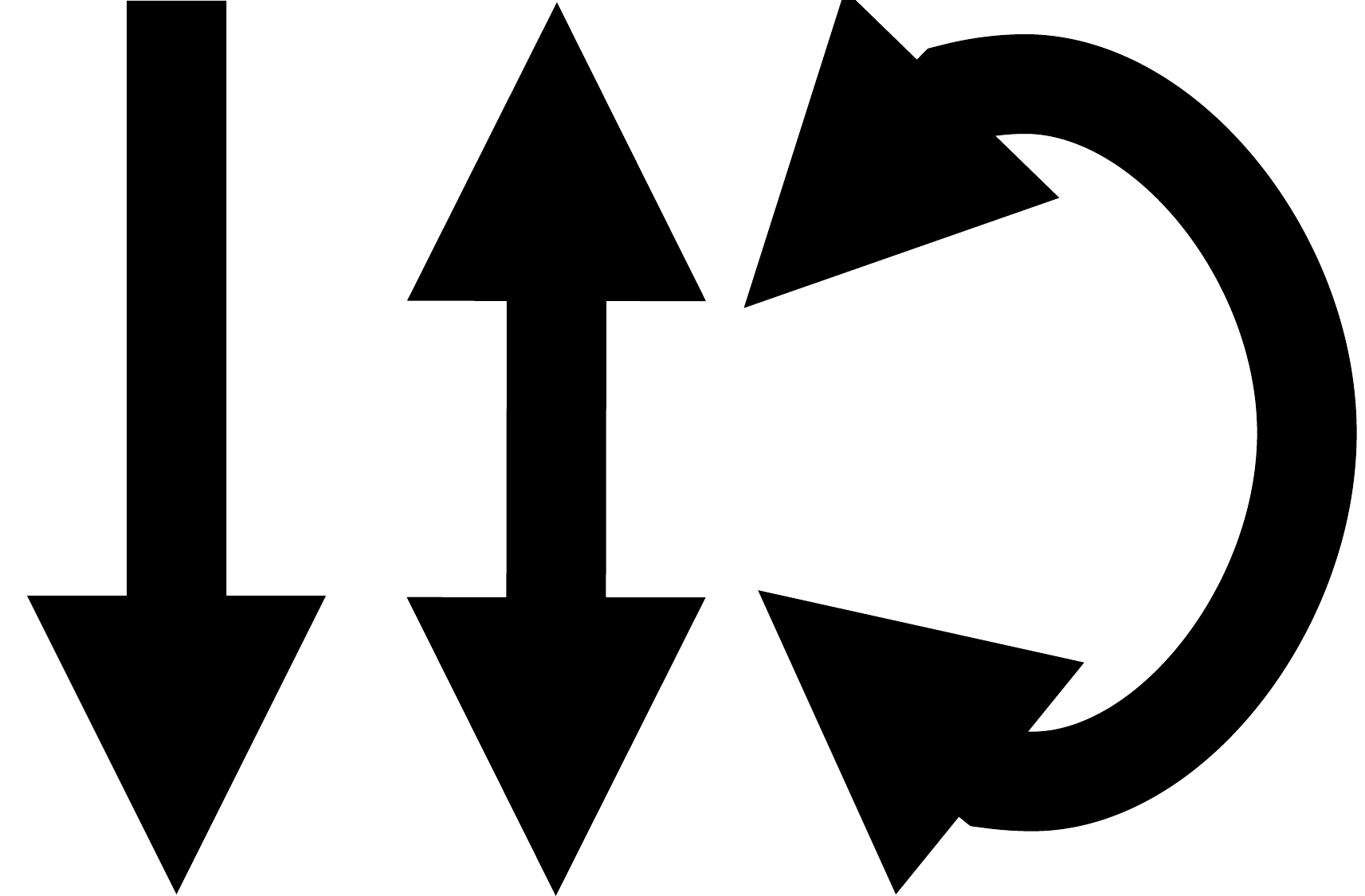}} Relationship.
        \ding{172} An internal relationship within a single entity
         (e.g., self-encouragement).
        \ding{173} A unidirectional relationship between two entities
         (e.g., entity 1 encourages entity 2).
        \ding{174} A bidirectional relationship between two entities
         (e.g., entity 1 and entity 2 encourage each other).
        \ding{175}, \ding{176} and \ding{177} Represent unidirectional relationships of ``single entity to group of multiple entities'', ``group of multiple entities to single entity'' and ``group of multiple entities to multiple entities'' respectively.
        \ding{178} Complex relationships between entities, either unidirectional or bidirectional, each relationship independently controlled.
        \ding{179} In TaleFrame, the story evolves in linear time and does not support causal reasoning or non-linear narratives.
  }
  \label{fig:interactionpattern}
\end{figure}

Volunteers (P1–P4) demonstrated a preference for story creation using a drag-and-drop card interface over text-based outlining methods, citing greater enjoyment with the former.
However, some volunteers also voiced concerns regarding the iterative process of editing unit attributes (P1, P4). P1 noted, ``\textit{When creating a story, my first instinct is always to focus on the broader framework, so I start by using event cards and entity cards to outline the story. Then, I fill in the attributes of the event and entity cards.}'' P4 added, ``\textit{I had to revise several times during the creation process. If the attribute editing could be made more convenient, I would be even more inclined to use this method.}'' Additionally, some volunteers suggested that certain semantic information within the story could be conveyed through interactive methods(P2, P3). P2 said, ``\textit{I prefer placing entity cards involved in the same event on top of the event cards. Attaching entity cards to event cards allows me to easily see which entities are involved in the event.}'' P3 proposed, ``\textit{Relationship cards could be used to connect different entity cards. If event cards were organized with connecting lines, it would more effectively illustrate the sequence and flow of events.}''
Building on these insights, we identify key interaction patterns (Figure~\ref{fig:interactionpattern}) and propose the following three design goals:

\begin{itemize}
    \item DG1: The system should allow interfaces for entering the attributes of each foundational unit to facilitate the user's construction of the initial story frame.
    \item DG2: The system should help users understand the overall narrative structure, facilitating the analysis of connections between entities and relations among events.
    \item DG3: The system should provide evaluations and targeted suggestions to improve the current story framework.
\end{itemize}

\section{TaleFrame: Methodology and Dataset Construction}
In this section, we will introduce the methodology for guiding LLMs in generating structured stories, introducing the preference dataset used to fine-tune the Llama-3-8B model~\cite{dubey2024llama}. The dataset was created by GPT-4o~\cite{chatgpt4o2024}. Details on the prompts, datasets, and model settings are in the supplemental materials.

\begin{table*}[htbp]
\caption{Components of the TIDD-EC Framework and Prompts Used in Entity Extraction}
\label{tab:t3}
\centering
\begin{tabular}{lp{4.5cm}p{10.5cm}}
\hline
Components         & \multicolumn{1}{c}{Explanation}
& \multicolumn{1}{c}{Detailed Prompt \textit{in Entity Extraction}} \\ 
\hline
T-Task             & Clearly indicates the type of activity the GPT-4o is expected to perform.
& \textit{Extract all entities information from the story, creating an ‘entities‘ array of objects. Each object should contain details about the name, identity, motivation\dots}\\
I-Instruction      & Summarize the specific steps or guidelines that LLMs should follow.              
& \textit{1. Read the entire story carefully, identifying all entities, including primary\dots}
\newline \textit{2. For each entity, create an `entity' object.}
\newline \textit{3. Ensure each `entity' object contains the entity’s name, identity, motivation\dots} \\
D-Do               & Specify the actions that the GPT-4o should take in their responses.                
& \textit{- Assign each entity a unique `entity\_id' with the format `entity\_1', `entity\_2' \dots}
\newline \textit{- Enter the entity's name in the `entity\_name' field.}
\newline \textit{- Describe the entity's identity in the `entity\_identity' field\dots}
\newline \textit{- Specify the entity's motivation in the `entity\_motivation' field \dots}
\newline \textit{- List at least one personality trait in the `personality\_traits' field\dots}
\newline \textit{- Include all entities with any form of contribution to the story\dots} \\
D-Don't            & Specify the actions that the GPT-4o should avoid in their responses.               
& \textit{- Do not exclude any entity, no matter how small their role in the story.}
\newline \textit{- Avoid vague descriptions; make sure content in each field is clear and specific.}
\newline \textit{- Do not omit any required fields (i.e., `entity\_id', `entity\_name', \dots).} \\ 
E-Example          & Provide specific examples of desired outcomes or responses.                      
& \textit{Story content: ``Young explorer named Alice is full of adventurous spirit \dots''}
\newline \textit{Example output: }\texttt{[\{ ``entity\_id'': ` '\dots ]}\\
C-Content          & Data provided by the user.                                                       
& \textit{Original Content: ``Sara and Ben were playing in the park. They saw \dots''}\\
\hline
\end{tabular}
\end{table*}

\subsection{Prompting Strategies for Parsing Tasks}
Prompt engineering is a technical methodology designed to optimize and control how pre-trained language models respond to input instructions.
Specifically, we employed \textbf{prompt chaining}\cite{wu2022ai,sun2024prompt}, which enhances the accuracy and reliability of processing outcomes in LLMs by decomposing complex tasks into consecutive sub-tasks. 
We decompose the task of story parsing into four sub-tasks: \textit{extract entities}, \textit{extract events}, \textit{extract story outline}, and \textit{extract relationships}. The prompts from preceding steps are provided to the LLMs, with the results incorporated as components of prompts for subsequent sub-tasks, as shown in Figure\ref{fig:pipeline}.b.

Prompt frameworks have been shown to optimize interactions with LLMs, ensuring the precise execution of instructions~\cite{co-star2024}. 
In this paper, we adopt the TIDD-EC prompt framework~\cite{tidd-ec2024}, which consists of five key components: \textit{task type}, \textit{detailed instructions}, \textit{dos and don'ts}, \textit{examples}, and \textit{user content}. 
By clearly defining the task type, providing detailed execution guidelines, outlining error-prevention measures, using specific examples to enhance understanding, and incorporating user-specific content, the TIDD-EC framework ensures the generation of more accurate and relevant responses. 
Table~\ref{tab:t3} presents the prompts used during the ``Entity Extraction'' task based on the TIDD-EC framework. 
Additionally, we experimented with four prompting strategies and compared their performance on the task of parsing stories into JSON format: 

\begin{itemize}
    \item \textbf{Zero-shot prompting:} Described the parsing task in natural language, without providing specific examples~\cite{radford2019language}.
    \item \textbf{TIDD-EC framework:} Framed the parsing task using the TIDD-EC framework~\cite{tidd-ec2024}.
    \item \textbf{TIDD-EC + CoT:} Built upon the second strategy by incorporating chain-of-thought (CoT) reasoning~\cite{wei2022chain}.
    \item \textbf{TIDD-EC + prompt chaining:} Divided the parsing task into four subtasks, each described using TIDD-EC, and linked these subtasks via prompt chaining~\cite{sun2024prompt}.
\end{itemize}

\begin{figure}[htbp]
    \centering
    \includegraphics[width=\linewidth]{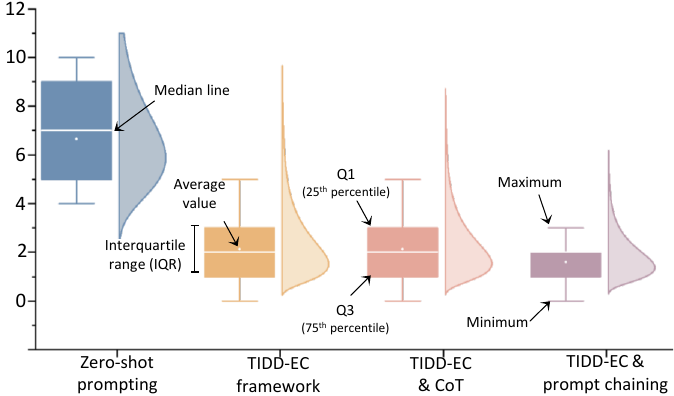}
    \caption{Comparison of Prompting Techniques: Box and violin plots show tree edit distances (lower is better) for Zero-shot prompting, TIDD-EC framework, TIDD-EC with CoT, and TIDD-EC with prompt chaining. }
    \label{fig:Prompt_strategy}
\end{figure}

The performance of the four strategies is shown in Figure~\ref{fig:Prompt_strategy}. It is evident that the combination of prompt chaining and TIDD-EC excels in the structured parsing of short stories. The details of the adopted prompt chaining with the TIDD-EC framework are presented in the supplemental materials.


\subsection{Dataset Construction}
We randomly selected a subset (about 9851 stories) from Tinystories~\cite{eldan2023tinystories}. The original stories were parsed into JSON format using prompt chaining and combined with the TIDD-EC. During the parsing process, elements such as events, entities and settings of the stories are extracted and structured into JSON format. Based on this structured data, GPT-4 will generate new stories that retain core themes and structure while allowing for innovation.

\begin{figure}[htbp]
  \centering
  \includegraphics[width=\linewidth]{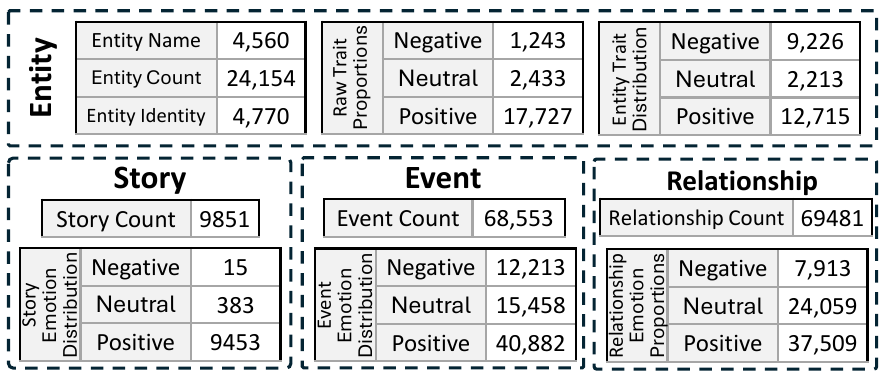}
  \caption{The statistical analysis of the preference dataset.
  }
  \label{fig:datad}
  \vspace{-1em}
\end{figure}

We purposed a set of evaluation criteria specifically tailored to assess the quality of stories, encompassing a comprehensive analysis across seven critical dimensions: \textit{Functionality}, \textit{Technicality}, \textit{Innovativeness}, \textit{Readability}, \textit{Thoughtfulness}, \textit{Emotional authenticity}, and \textit{Clarity of perspective}. 
Based on this criteria, third-party LLMs (such as GPT-4o) will evaluate both the original and revised versions of the stories, assigning them ``Chosen'' or ``Rejected'' labels.
Furthermore, the structured information of the stories will be recorded in JSON format and, together with the labeled stories, will form the preference dataset. 
The emotional distributions and other features in the preference dataset, are illustrated in Figure~\ref{fig:datad}.

\subsection{Model Fine-tuning}
The dataset was initially divided into a training set and a test set at a ratio of 9:1. The training set contains 8,866 preference pairs utilized for model fine-tuning, while the test set contains 985 preference pairs that remain unseen by the model.
Subsequently, we fine-tuned Llama-3-8B model with the AdamW optimizer. 
We used a LoRA-based fine-tuning method with a set learning rate of 5e-6, a batch size of 10, and a training epoch of 10, resulting in an evaluation loss of 0.038 on the test set. The selection of these parameters was informed by model documentation recommendations, trial-and-error, and the computational resources available.

Furthermore, we conducted an ablation study methodology to assess the importance of different compositional units in story generation tasks. A baseline model was constructed based on the Llama-3-8B, integrating all the involved units to produce coherent and controlled stories. In the ablation study, different categories of units were sequentially removed, such as the ``event'' unit, and the model was retrained and fine-tuned to adapt to the modified data structure. All models were trained under consistent conditions, including fixed initialization parameters, learning rate, batch size, and training epochs, to ensure the fairness and comparability of the experiments. The performance of each ablated model was evaluated on an independent test set. Additionally, the quality of the stories was quantitatively scored from seven dimensions using same evaluation criteria based on GPT-4o. The results of these evaluations will be detailed in Section 6.

\section{TaleFrame Interface}
\begin{figure}[htbp]
  \centering
  \begin{minipage}[t]{0.48\linewidth}
    \centering
    \includegraphics[width=0.85\textwidth]{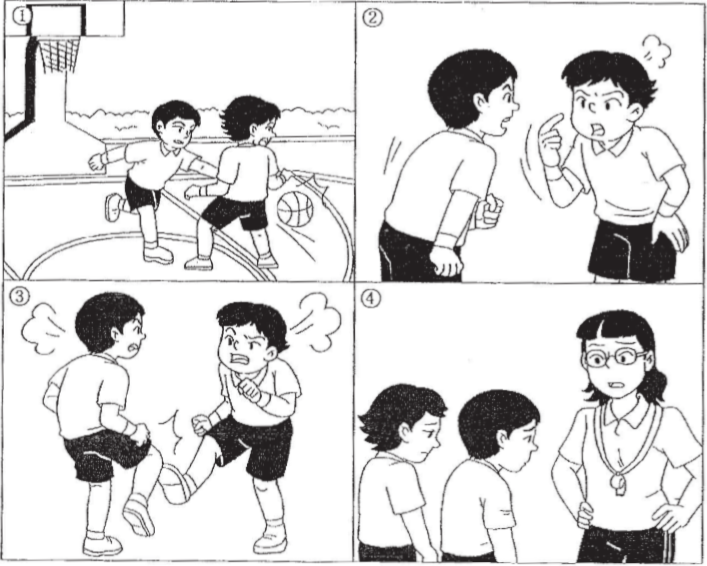}
    \centerline{(a)}
  \end{minipage}
  \begin{minipage}[t]{0.48\linewidth}
    \centering
    \includegraphics[width=\textwidth]{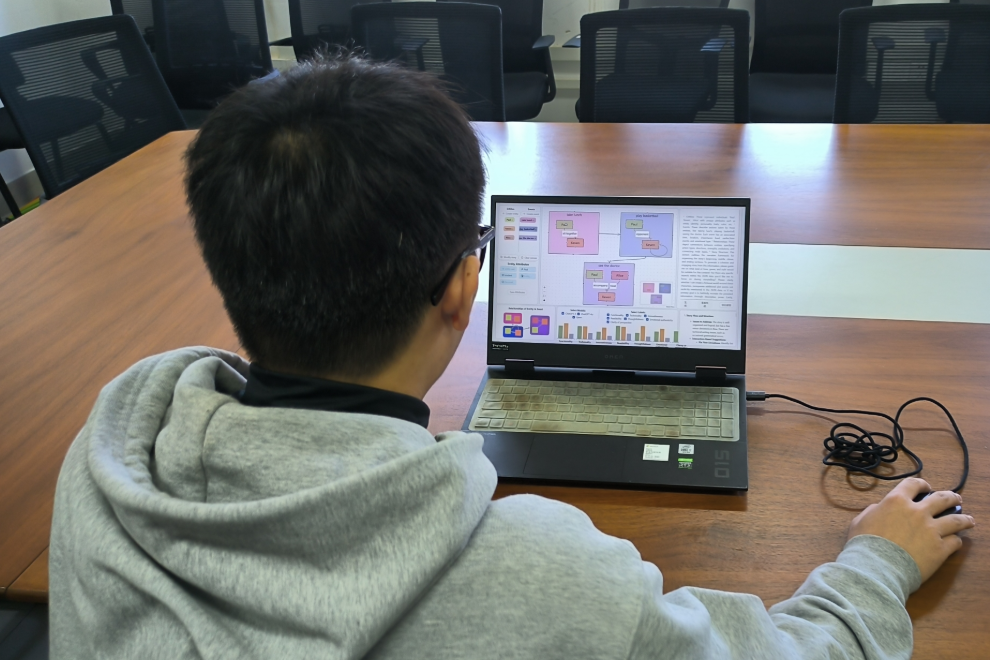}
    \centerline{(b)}
  \end{minipage} 
  \caption{An example of story creation using TaleFrame. (a) This diagram has three entities, four events, with clear relationships and a linear evolution of events, which is ideal as an introduction to how to use the TaleFrame. (b) An on-site photo of using TaleFrame.}
  \label{fig:picture_composition}
\end{figure}

In this section, we illustrate the features of our interface using a use case scenario of ``Picture Composition''. Picture composition offers users obvious information, such as the number of entities and events, and types of relationships, providing an overview to guide the creation process.
Figure~\ref{fig:picture_composition}.a illustrates a diagram for picture composition, while Figure~\ref{fig:picture_composition}.b illustrates the story creation process of P4 with TaleFrame.


\begin{figure*}[htbp]
  \centering
  \includegraphics[width=\linewidth]{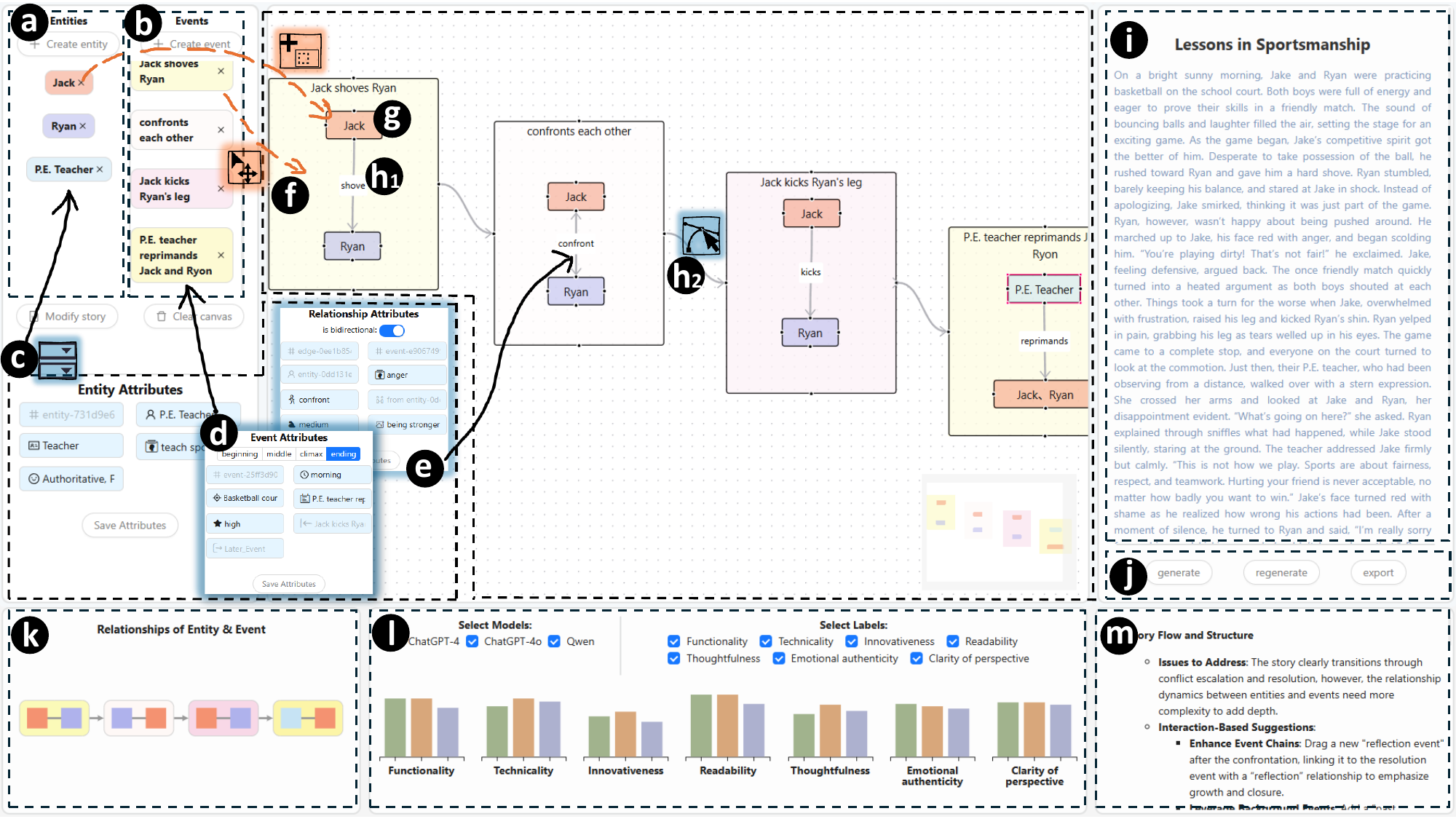}
  \caption{The interface of TaleFrame is shown with a usage scenario. TaleFrame consists of three distinct views: interactive view (a-h), textual view (i-j), and detail view (k-m). The interactive view facilitates the construction of a story framework through the creation of four foundational units via four interaction modalities. The textual view enables users to browse and regenerate the story. The detail view provides story information, evaluates the current story across seven dimensions, and offers suggestions for modifications.}
  \label{fig:interface}
\end{figure*}

$\lozenge$ \textbf{Step 1: Create entity and event templates.}
In Step 1, 
P4 proceeds by creating three entities (Figure~\ref{fig:interface}.a), followed by the assignment of names and identities to each entity (Figure~\ref{fig:interface}.c) (e.g., \textit{entity\_name: Jack} and \textit{entity\_identity: student}).
The attributes of the entities will be synchronized to the entity templates after clicking the ``Save Attributes'' button \raisebox{-0.7mm}{\includegraphics[scale=0.03]{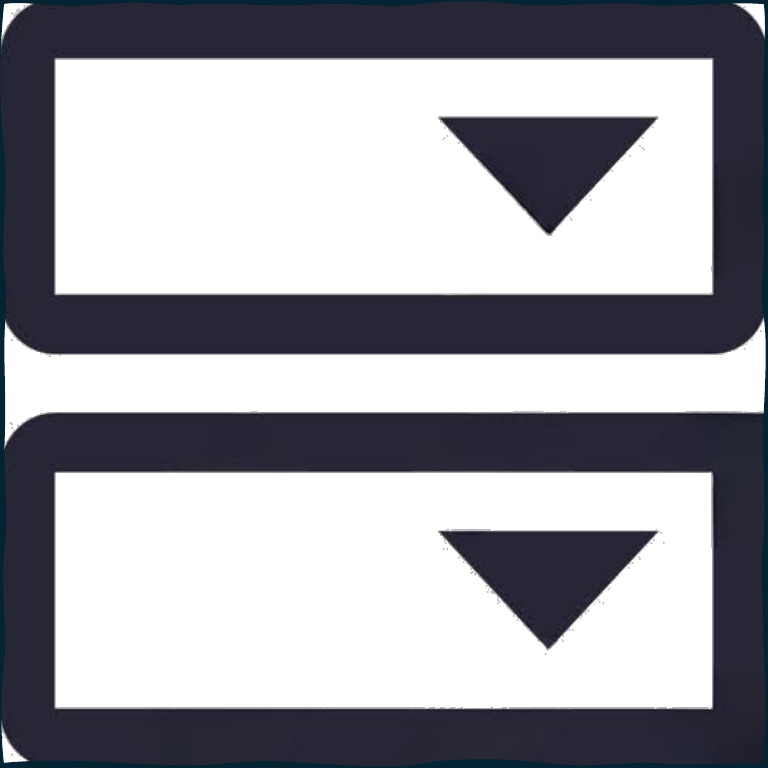}}.
Next, he creates four events (Figure~\ref{fig:interface}.b) and edits the time and location for each event (Figure~\ref{fig:interface}.e) (e.g., \textit{event\_time: morning; event\_location: basketball court}) \raisebox{-0.7mm}{\includegraphics[scale=0.03]{pictures/edit_attributes.pdf}}.

$\lozenge$ \textbf{Step 2: Interactively build a story frame.}
In Step 2, P4 begins the iterative process of refining the templates and organizing the story framework through drag-and-drop \raisebox{-0.7mm}{\includegraphics[scale=0.03]{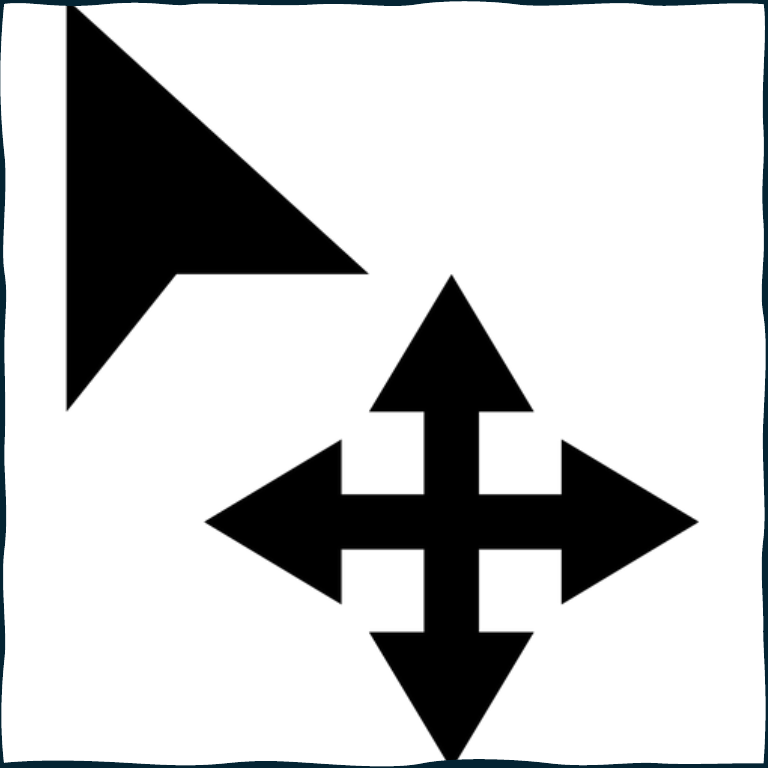}}.
P4 starts by selecting the first event template, adding event details (e.g., \textit{event\_details: Jack shoves Ryan}), and clicks the ``Save Attributes'' button to synchronize these information with the first event template (Figure~\ref{fig:interface}.d) \raisebox{-0.7mm}{\includegraphics[scale=0.03]{pictures/edit_attributes.pdf}}. 
P4 drags and drops the event ``Jack shoves Ryan'', which represents the first grid in the diagram, onto the canvas (Figure~\ref{fig:interface}.f) \raisebox{-0.7mm}{\includegraphics[scale=0.03]{pictures/drag_drop.pdf}}. Next, P4 attaches the two entities involved in this event to the box (Figure~\ref{fig:interface}.g) \raisebox{-0.7mm}{\includegraphics[scale=0.03]{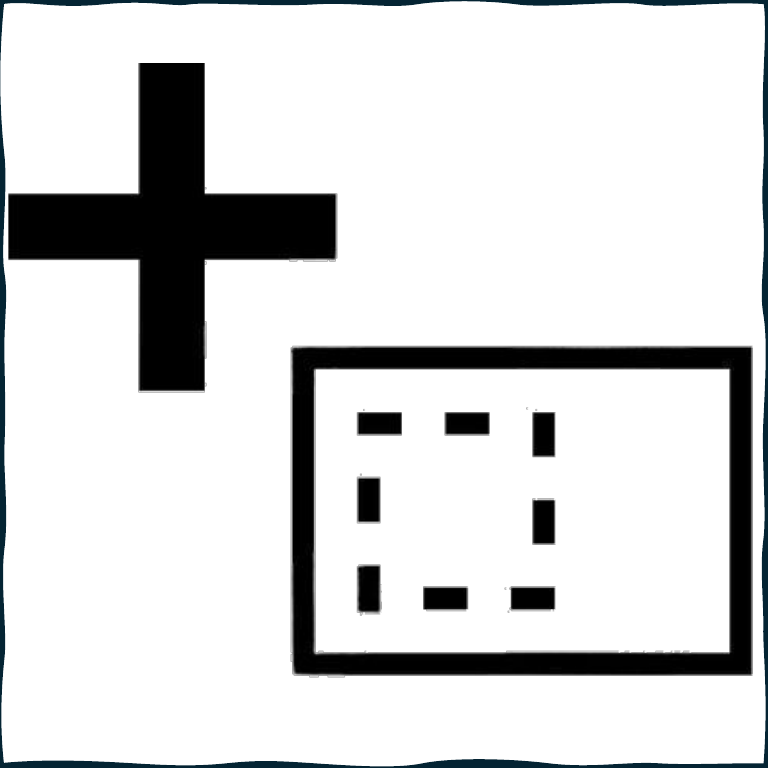}}.
Then, P4 connects the two entities (Figure~\ref{fig:interface}.h1) \raisebox{-0.7mm}{\includegraphics[scale=0.03]{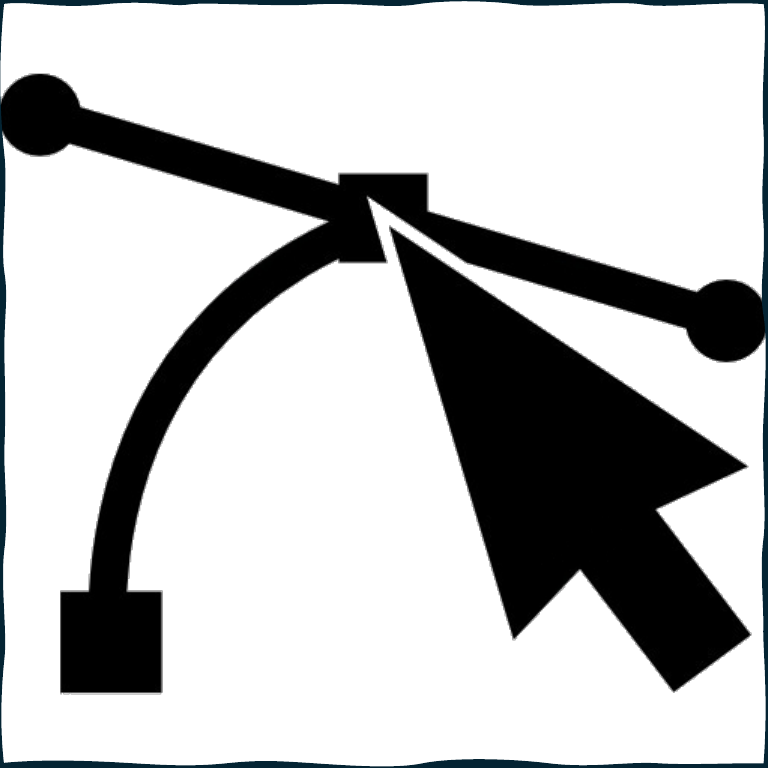}}, clicks on the connection line, and edits the relationship attributes (Figure~\ref{fig:interface}.e) \raisebox{-0.7mm}{\includegraphics[scale=0.03]{pictures/edit_attributes.pdf}} (e.g., \textit{emotion\_type: intense}; \textit{action\_type: shove}; \textit{relationship\_strength: low}; \textit{relationship\_evolution: being stronger}). 
The ``action\_direction'' attribute of the relationship is represented by the ``connect node'', which defaults to the start point and end point (e.g., ``from Jack to Ryan''). However, for the event ``confront each other'' in the second grid, where the action ``confront'' is mutual, a switch button labeled ``is bidirectional'' \raisebox{-0.7mm}{\includegraphics[scale=0.03]{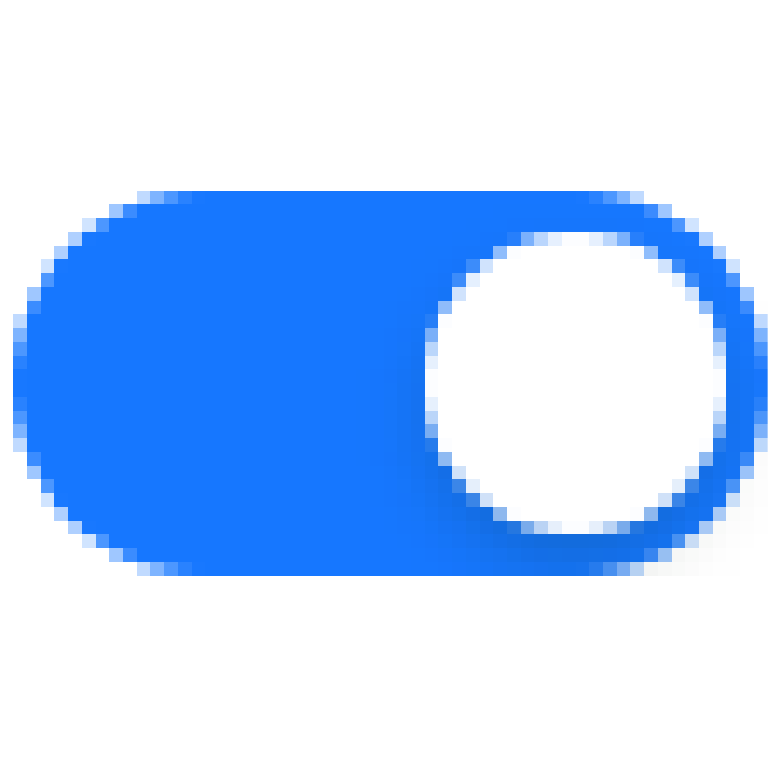}} is provided. When activated, this switch automatically sets the action\_direction to ``bidirectional''.
By repeating above process, P4 completes the remaining three grids, linking the four event boxes to finalize the entire story framework (Figure~\ref{fig:interface}.h2) \raisebox{-0.7mm}{\includegraphics[scale=0.03]{pictures/connect_node.pdf}}. Finally, P4 clicks the ``Modify Story'' button, edits the title and description, and assigns the four events to the appropriate stages of the story.

$\lozenge$ \textbf{Step 3: Feedback iterative revision of the story.}
In Step 3, upon clicking the ``Generate'' button (Figure~\ref{fig:interface}.j), P4 triggers TaleFrame to automatically convert the constructed story framework into a structured JSON format and transmit it to the back-end fine-tuned model.
The model then generates the story and presents it in the textual view (Figure~\ref{fig:interface}.i). Concurrently, in the detail view, TaleFrame displays the relationships of entities and events (Figure~\ref{fig:interface}.k), providing an overview of the story.
Using three different third-party LLMs as evaluators, the story is scored on seven dimensions according to the same evaluation criteria (Figure~\ref{fig:interface}.l). 
And a comprehensive suggestion will also be generated based on the feedback (Figure~\ref{fig:interface}.m). 
According to the suggestion, P4 modifies aspects with lower scores (e.g., adding relationships to certain events). After that, P4 clicks the ``Regenerate'' button to generate a new version. Finally, P4 clicks the ``Export'' button to obtain the final story, the framework diagram, and the JSON file.

\section{Evaluation}

\begin{table*}[htbp]
\caption{Results of the ablation study evaluating the impact of different units on model performance.}
\label{tab:tb5}
\centering
\resizebox{0.98\linewidth}{!}{
\begin{tblr}{
    colspec = {|c|c|c|c|c|c|c|c|c|},
    vline{1,2,3,4,5,6,7,8,9,10} = {0.3mm},
}
\hline[0.3mm]
\SetCell[c=2]{c} \diagbox[width=4.5cm]{Fine-tuned models}{Metrics}  & & \textbf{Functionality} & \textbf{Technical} & \textbf{Innovative} &
 \textbf{Readability} & \textbf{Thoughtfulness} & \textbf{Emotional} & \textbf{Clarity} \\
\hline[0.3mm]
\SetCell[r=1]{c} \textbf{Full Unit}s & Generated & 
\SetCell[c=1]{c, mycolor1} \textbf{3.91} & 
\SetCell[c=1]{c, mycolor1} \textbf{3.95} & 
3.20 & 
\SetCell[c=1]{c, mycolor1} \textbf{3.97} & 
\SetCell[c=1]{c, mycolor1} \textbf{3.54} & 
\SetCell[c=1]{c, mycolor1} \textbf{3.77} & 
\SetCell[c=1]{c, mycolor1} \textbf{3.88} \\
\hline[0.3mm]
\SetCell[r=2]{c} {\textbf{Without} \\ \textbf{Events}} & Generated & 
3.85 & 3.92 & 
\SetCell[c=1]{c, mycolor1} \textbf{3.23} & 
3.88 & 3.52 & 3.75 & 3.86 \\
\hline[0.3mm]
 & U-test & 0.334 & 0.786 & 0.615 & 
 \SetCell[c=1]{c, mycolor2} \textbf{0.029} & 
 0.667 & 0.310 & 0.814 \\
\hline[0.3mm]
\SetCell[r=2]{c} {\textbf{Without} \\ \textbf{Relationships}} & Generated & 
3.83 & 3.88 & 3.15 & 3.89 & 
3.44 & 3.66 & 3.80 \\
\hline[0.3mm]
& U-test & 
\SetCell[c=1]{c, mycolor2} \textbf{0.046} & 0.119 & 0.330 & 
\SetCell[c=1]{c, mycolor2} \textbf{0.011} & 
\SetCell[c=1]{c, mycolor2} \textbf{0.017} & 
\SetCell[c=1]{c, mycolor2} \textbf{0.022} & 
\SetCell[c=1]{c, mycolor2} \textbf{0.014} \\
\hline[0.3mm]
\SetCell[r=2]{c} {\textbf{Without} \\ \textbf{Entities}} & Generated & 3.84 & 3.87 & 3.13 & 3.88 & 3.44 & 3.68 & 3.81 \\
\hline[0.3mm]
 & U-test & 0.085 & 0.083 & 0.176 & 
 \SetCell[c=1]{c, mycolor2} \textbf{0.003} & 
 \SetCell[c=1]{c, mycolor2} \textbf{0.036} & 
 0.338 & 
 \SetCell[c=1]{c, mycolor2} \textbf{0.031} \\
\hline[0.3mm]
\SetCell[r=2]{c} {\textbf{Without} \\ \textbf{Outline}} & Generated & 3.86 & 3.91 & 3.17 & 3.91 & 3.46 & 3.71 & 3.85 \\
\hline[0.3mm]
 & U-test & 0.186 & 0.796 & 0.534 & 
 \SetCell[c=1]{c, mycolor2} \textbf{0.012} & 
 0.077 & 0.524 & 0.397 \\
 \hline[0.3mm]

\end{tblr}
}
\end{table*}

In the ablation study, the experimental platform was configured as follows: Ubuntu operating system, a 13th Gen Intel Core i9-13900K processor, 128GB of memory, and an NVIDIA RTX A6000 GPU. 
In the usability study of the TaleFrame interface, we conducted user sessions with the same volunteers who were mentioned in Section 3. We also developed a multi-dimensional user experience evaluation questionnaire, which
focused on aspects such as functionality, technical performance, and innovation.

\subsection{Ablation Study}
To gain a deeper understanding of the four foundational units' role in controllable story generation, we conducted an ablation study to analyze the impact of each unit on the generated output. In the ablation study, we fine-tuned and trained five distinct models based on the same Llama-3-8B model. The training datasets included the following preference datasets: one with the complete set of foundational units, one without events, one without relationships, one without entities, and one without the story outline.

We evaluated both qualitative and quantitative metrics. Specifically, the same test dataset (comprising 196 samples) was used to assess the fine-tuned models. Qualitative metrics were employed to score and assess various aspects of the generated stories, such as functionality and artistic quality. We used the stories chosen in the preference dataset (``Chosen'') as a baseline for comparison, enabling a quantitative analysis of the similarity between the stories generated by different fine-tuned models. Thus, similarity metrics were utilized to evaluate the impact of the four foundational units.

\textbf{Qualitative Evaluation}: Qualitative indicators are based on the seven evaluation dimensions mentioned in Section 3.1. A 5-point Likert scale (1-5) was developed
to assess the model’s performance. 
The same evaluation prompt was used across all assessments, and a third-party LLMs (GPT-4o) conducted the evaluations. 
Each test sample underwent three independent evaluations, with the average score serving as the final result, as shown in the ``Generated'' row of Table~\ref{tab:tb5}. 
``Full Units'' refers to the complete model, while other rows show the results after omitting specific units, including Events, Relationships, Entities, and Outline. 
The significant differences, determined by U-tests ($p < 0.05$), are highlighted in green, and the best scores for each metric are marked in yellow.

The result shows that the model fine-tuned with the complete set of four foundational units performs well across most aspects, including functionality, technicality, readability, thoughtfulness, and emotional clarity. It exhibits the most significant gap in readability. However, due to the constraints, it does not perform the best in terms of creativity. 
In order to further validate the significant differences between different fine-tuned models, we conducted a Mann-Whitney U test on the ratings of the seven dimensions of the generated stories, as shown in the ``U-test'' row of Table~\ref{tab:tb5}.
The U-test results reveal the statistical significance of differences in various evaluation metrics when specific units are removed. Notably, the removal of the ``Relationships'' unit affected several metrics, including functionality ($p = 0.046$), readability ($p = 0.011$), thoughtfulness ($p = 0.017$), emotionality ($p = 0.022$), and clarity of perspective ($p = 0.014$), highlighting its crucial role in maintaining the quality of generated outputs. Similarly, the removal of the ``Entities'' unit impacted readability ($p = 0.003$), thoughtfulness ($p = 0.036$), and clarity of perspective ($p = 0.031$). In contrast, the removal of the ``Events'' and ``Outline'' units showed differences only in the readability dimension ($p = 0.029$, $p = 0.012$). These results suggest that among the four core units, ``Relationships'' and ``Entities'' have the greatest overall impact on the quality of the generated story, while the ``Events'' and ``Outline'' units primarily affect readability.

\begin{figure}[htbp]
    \centering
    \includegraphics[width=\linewidth]{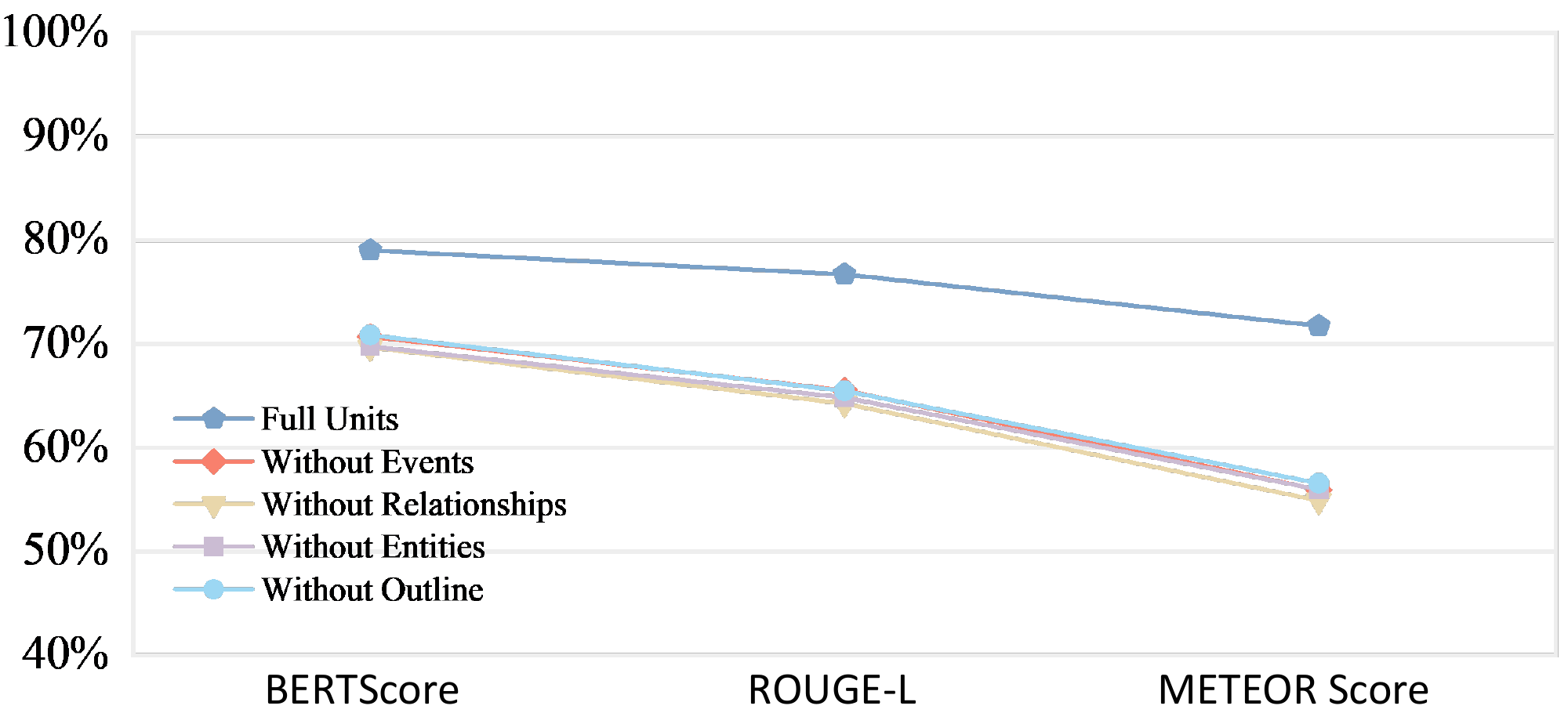}
    \caption{The ablation study results illustrate the performance of different configurations (Full Units, Without Events, Without Relationships, Without Entities, Without Outline) relative to the baseline preference dataset (Chosen) across three evaluation metrics: BERTScore, ROUGE-L, and METEOR Score.}
    \label{fig:evaluation}
\end{figure}

\textbf{Quantitative Evaluation}: 
We primarily compared the similarity between the content generated by different fine-tuned models, using the same JSON input, and the original text. Each fine-tuned model generates three stories for each JSON. We used the original text as the benchmark and assessed similarity using three metrics: ROUGE-L, BERTScore, and METEOR Score. ROUGE-L measures the similarity between two sequences based on the length of their longest common subsequence (LCS). BERTScore captures semantic similarity, avoiding issues related to surface-level matching. METEOR Score combines various matching techniques, offering greater flexibility and robustness.
As shown in Figure~\ref{fig:evaluation}, the performance of different configurations 
relative to the preference dataset baseline is presented. 
The Full Units configuration performs best in terms of content control, semantic consistency, and linguistic naturalness in story generation.

\subsection{Usability Study}
\textbf{Participates \& Task}. 
We re-invited the 2 experts and 17 volunteers mentioned in Section 3 to complete the TaleFrame usability study.
To minimize the inconvenience caused by a lack of inspiration during the creative process, we provided volunteers with various of pictures. Volunteers were allowed to choose topics based on their preferences and engage with TaleFrame to generate stories. During the creative process, if volunteers were dissatisfied with the generated story, they could modify the attributes of the four foundational units in order to create a story that met their expectations. Overall, volunteers were required to engage with all of the interactive features provided by the TaleFrame.

\textbf{Procedure}.
The experiment lasted approximately 20 to 35 minutes. Initially, volunteers were asked to select a picture of their interest for a writing task. We ensured that volunteers fully understood the content of the selected picture. Following this, volunteers were introduced to the features and interactions of TaleFrame. Then, volunteers were asked to create a story based on their selected pictures. Throughout the experiment, we recorded the inputs and actions of the volunteers. We also conducted interviews to collect feedback, when the task was completed. Each volunteer received a compensation of \$5.

\begin{figure}[htbp]
    \centering
    \includegraphics[width=\linewidth]{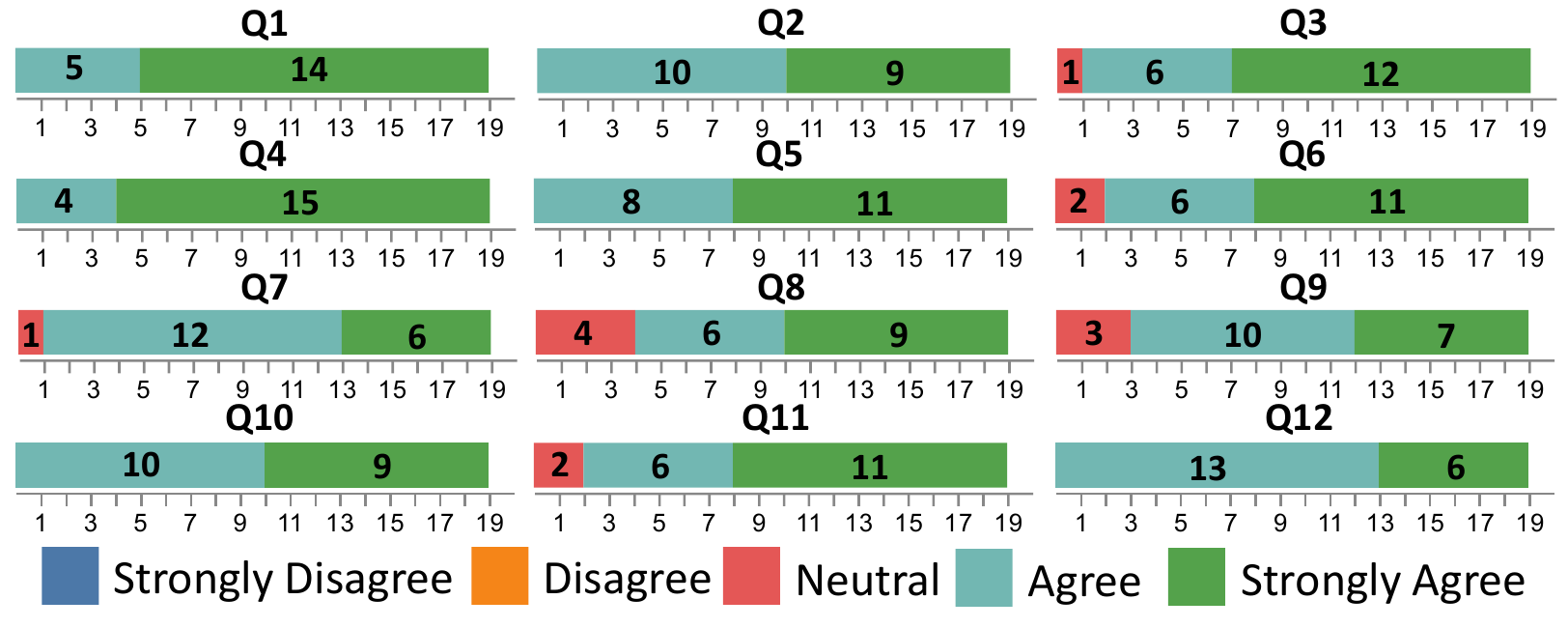}
    \caption{User feedback ratings for each dimension of TaleFrame. This figure shows the distribution of volunteer ratings for each dimension of TaleFrame, based on a 5-point Likert scale: Framework Construction (Q1-Q3), Narrative Visualization (Q4-Q6), Narrative timing and coherence (Q7-Q9), and Evaluation and Improvement Suggestions (Q10-Q12).}
    \label{fig:user_study}
\end{figure}

\textbf{User Feedback.}
Volunteers evaluated TaleFrame using a 5-point Likert scale based on their user experience, as shown in Figure~\ref{fig:user_study}.
\ding{172} Framework Construction (Q1-Q3): TaleFrame provides users with an efficient and convenient tool to construct a comprehensive and structured story framework. This is facilitated by an intuitive user interface and input options, allowing users to easily define entities, events, and relationships within the story (DG1).
\ding{173} Narrative Visualization (Q4-Q6): TaleFrame supports users in intuitively constructing and arranging entities, events, and relationships through interactive methods. It also clearly visualizes the connections between these elements, making the story structure easy to understand. Furthermore, flexible tools are provided to help users adjust or optimize the story structure during the creation process (DG1, DG2).
\ding{174} Narrative timing and coherence (Q7-Q9): TaleFrame offers comprehensive support to ensure the temporal coherence of the story. It helps users analyze the sequence of events, maintain consistency in the progression of entity behaviors over time, and identify potential temporal flaws in the story structure. In addition, it provides specific recommendations for improving emotional expression and thematic depth, supporting the creation of plot events or entity relationships that better align with the timeline (DG2, DG3).
\ding{175} Evaluation and Improvement Suggestions (Q10-Q12): TaleFrame is capable of accurately evaluating the quality of the current story framework and offering actionable, targeted optimization suggestions. This enables users to enhance the overall quality of the story (DG3).

Most volunteers had positive opinions regarding TaleFrame's performance in constructing the interactive framework of the system (Q1: $\mu=4.74$, $\sigma=0.44$; Q2: $\mu=4.47$, $\sigma=0.50$; Q3: $\mu=4.58$, $\sigma=0.59$). Most volunteers found TaleFrame's story outline visualization to be effective. In Q4, they reported that the drag-and-drop feature was intuitive for creating and organizing event and entity templates (Q4: $\mu=4.74$, $\sigma=0.41$). Although most volunteers found TaleFrame clear and helpful in showing how templates connect (Q5: $\mu=4.58$, $\sigma=0.49$), two neutral ratings in Q6 suggested that providing some default templates could further enhance the system's usability (Q6: $\mu=4.47$, $\sigma=0.68$). Volunteers gave positive feedback on the consistency of event sequences (Q7: $\mu=4.26$, $\sigma=0.55$) and actions/progression (Q8: $\mu=4.26$, $\sigma=0.78$. However, their performance in analyzing temporal consistency and detecting issues in complex narratives was below expectations (Q9: $\mu=4.2$, $\sigma=0.68$), indicating that the system's support for handling complex storylines might need further improvement. Volunteers found TaleFrame assessment to be accurate and valuable (Q10: $\mu=4.47$, $\sigma=0.499$), although some felt that the improvement suggestions were not intuitive and difficult to apply (Q11: $\mu=4.47$, $\sigma=0.678$). However, they acknowledged that too much automation might not be ideal and, in general, they believed that TaleFrame could improve story quality (Q12: $\mu=4.32$, $\sigma=0.465$).

\section{Limitation and Future Work}
In this section, we discuss the limitations of TaleFrame on interaction modes and advanced narrative, as well as directions for future research. 

\begin{itemize}
    \item \textbf{Interaction and visualization.}
    TaleFrame currently operates on desktop and laptop environments, leveraging a combination of mouse and keyboard for precise control, which is crucial in connecting entities to identify relationships and linking events to recognize event sequences. User studies indicate a preference for drag-and-drop interactions; however, due to the smaller screen sizes of popular mobile devices (e.g., smartphones and tablets), this interaction mode may be less effective. Although touch pens and fingers can replace the mouse, the complexity of long interaction processes and input of relationship information can lead to accidental touches, thus impairing the user experience on mobile devices. To reduce cognitive load, we have optimized information visualization by color coding, icons, and connecting lines, and we have implemented adaptive layout resizing to accommodate different screen sizes. In future work, we plan to explore and define a broader range of interaction styles, including touchscreens, keyboard input, and voice commands, to create a flexible framework that allows for diverse user engagement across various devices. 
    
    \item \textbf{Challenges in handling advanced narrative structures in story generation.}
    TaleFrame does not support handling complex event relationships beyond linear temporal control, especially causal relationships, flashbacks, and other narrative structures. In addition, the relationships between entities complicate the connections between events, impacting the coherence and accuracy of the generation process. While we have attempted to improve the modeling of event dependencies, the results have been less than ideal when handling multi-layered event interweaving, temporal reversals, and interactions between different entities. The relationships between entities, especially in events involving multiple entities, often lead to difficulties in generating the correct event sequence or capturing causal chains, which can result in incoherent or logically unclear storylines. As the complexity of the story plot and the relationships between entities increases, the model faces greater challenges in understanding and generating these event relationships. In future work, we will gradually attempt to improve entity relationship modeling and optimize strategies for generating non-linear narrative structures to enhance the model's performance in complex story generation, ultimately providing users with a richer and more coherent narrative experience.
\end{itemize}

\section{Conclusion}
In this paper, we present TaleFrame, an interactive story generation system that enhances both controllability and coherence. Using a fine-tuned Llama-3-8B model with prompt engineering, TaleFrame enables users to interactively adjust and customize story units, ensuring that the generated stories align with their creative intentions. TaleFrame effectively addresses the challenges of uncontrollable output from LLMs while maintaining privacy and personalization. Our approach demonstrates clear advantages in improving the precision and quality of story generation. This work offers a practical solution to interactive story creation and highlights the potential of fine-tuned language models with structured input to significantly enhance the control and quality of generated content.

\section*{Acknowledgments}
This work was supported by the National Key Research and Development Program of China (2022YFB3104800), National Natural Science Foundation of China (62422607, 62372411) and Zhejiang Provincial Natural Science Foundation of China (LR23F020003). Yunchao Wang is corresponding author.


%
\bibliographystyle{IEEEtran}
\bibliography{TaleFrame}

\vfill

\end{document}